\documentclass[10pt,twocolumn,letterpaper]{article}

\usepackage{cvpr}
\usepackage{times}
\usepackage{epsfig}
\usepackage{graphicx}
\usepackage{amsmath}
\usepackage{amssymb}
\usepackage{algorithm}
\usepackage{algorithmicx}
\usepackage{algpseudocode}
\usepackage{authblk}
\makeatletter
\renewcommand\AB@affilsepx{     \protect\Affilfont}
\makeatother

\newcommand{\algorithmicinit}{\textbf{}}
\newcommand{\INIT}{\item[\algorithmicinit]}
\algdef{SE}[DOWHILE]{Do}{doWhile}{\algorithmicdo}[1]{\algorithmicwhile\ #1}

\newcommand{\enc}{\mbox{Enc}}
\newcommand{\dec}{\mbox{Dec}}
\newcommand{\gen}{\mbox{Gen}}
\newcommand{\dis}{\mbox{Dis}}
\newcommand{\ali}{\mbox{Ali}}
\newcommand{\smo}{\mbox{smo}}
\newcommand{\pos}{\mbox{pos}}

\newcommand{\gan}{\mbox{gan}}
\newcommand{\vae}{\mbox{vae}}
\newcommand{\recons}{\mbox{recons}}
\newcommand{\loss}{\mathcal{L}}

\usepackage[breaklinks=true,bookmarks=false]{hyperref}

\cvprfinalcopy 

\def\cvprPaperID{0210} 

\begin{document}

\title{Crossing Nets: Combining GANs and VAEs with a \\Shared Latent Space for Hand Pose Estimation}

\author[1]{Chengde Wan}
\author[1]{Thomas Probst}
\author[1,3]{Luc Van Gool}
\author[2]{Angela Yao}
\affil[1]{ETH Z\"urich}
\affil[2]{University of Bonn}
\affil[3]{KU Leuven}

\maketitle

\begin{abstract}
State-of-the-art methods for 3D hand pose estimation from depth images require large amounts of annotated training data. We propose to model the statistical relationships of 3D hand poses and corresponding depth images using two deep generative models with a shared latent space. By design, our architecture allows for learning from unlabeled image data in a semi-supervised manner. Assuming a one-to-one mapping between a pose and a depth map, any given point in the shared latent space can be projected into both a hand pose 
and a corresponding depth map. 
Regressing the hand pose can then be done by learning a discriminator to estimate the posterior of the latent pose given some depth maps. To improve generalization and to better exploit unlabeled depth maps, we jointly train a generator and a discriminator. At each iteration, the generator is updated with the back-propagated gradient from the discriminator to synthesize realistic 
depth maps of the articulated hand, while the discriminator benefits from an augmented training set of synthesized and unlabeled samples.  The proposed discriminator network architecture is highly efficient and runs at 90\.FPS on the CPU with accuracies comparable or better than state-of-art on 3 publicly available benchmarks.
\end{abstract}


\section{Introduction}

\begin{figure}[t]
    \centering
    \includegraphics[width=\linewidth]{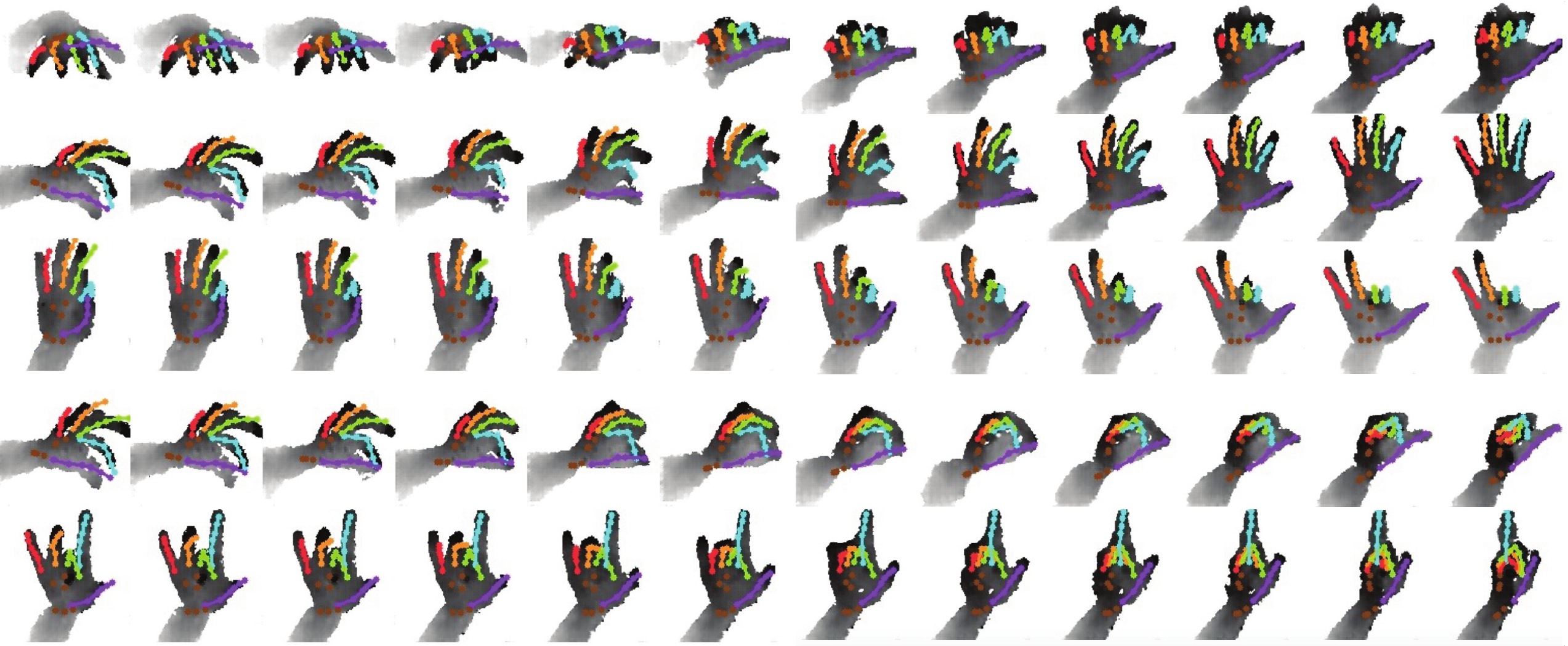}
    \caption{{\bf Random walk in the learned shared latent space.} A set of points is sampled on the connecting line between two points in the shared latent space. The pose and corresponding depth map are then reconstructed through our network. Our method generates meaningful and realistic interpolations in both pose and appearance space. 
    }
    \label{fig:render}
\end{figure}

We address the problem of estimating 3D hand pose from single depth images.  Accurate estimation of the 3D pose in real-time 
has many challenges, including the presence of local self-similarity and self-occlusions.
Since the availability of low-cost depth sensors, the progress made in developing fast and accurate hand trackers have relied heavily on having a large corpus of depth images annotated with hand joints. 
This is especially true for the recent success of deep learning-based methods~\cite{nyu_hand,Oberweger_15deep,Oberweger_15feedback,deepHand,Ge_multiview,handSA,Tekin2016,Vodopivec2016} which are all fully-supervised.

Accurately annotating 3D hand joints on a depth map is both difficult and time-consuming. While it is possible to synthesize data with physical renderers, there are usually discrepancies between the real and synthesized data.  Generated hand poses 
are not always natural nor reflective of poses seen in real-life applications. More importantly, it is very difficult to accurately model and render depth sensor noise in a realistic way. 

On the other ``hand'', it is very simple to collect an unlabeled dataset of real human hands with standard consumer depth cameras. This begs the question: how can one use these unlabeled samples for training?  To date, there has been virtually no work presented on semi-supervised learning for hand pose estimation. The one notable exception~\cite{Tang2013-sf} is a discriminative approach using transductive random forests and largely ignores high-order pixel correlations of unlabeled depth maps.

Previous works from neuroscience~\cite{synergy_neuroscience}, robotics~\cite{synergy_robotics} and hand motion capture~\cite{synergy_hand} have demonstrated that hand movements exhibit strong correlations between joints. We therefore conclude that the space of realistic hand poses
can be represented by a manifold in a lower-dimensional subspace. 
We further intuit that depth maps of the hand can be similarly encoded in a low-dimensional manifold, and be faithfully reconstructed with an appropriate generator.

In this paper, we propose a dual generative model that captures the latent spaces of hand poses and corresponding depth images for estimating 3D hand pose. We use 
the variational autoencoder\,(VAE) and the generative adversarial network\,(GAN) for modelling the generation process of hand poses and depth maps respectively. We assume a one-to-one mapping between a depth map and a hand pose; in this way, one can consider the latent hand pose space and latent depth map space to be shared.  Having a shared space is highly beneficial, since a point sampled in either latent space can be expressed both as a 3D pose, via the VAE's decoder, or as a depth map, through the GAN's generator. 

Fig.~\ref{fig:framework} gives an overview of our proposed framework. Our core idea is to learn a bi-directional mapping that relates the two latent spaces of hand poses and depth maps and therefore link together the pose encoder network with the generative models for hand poses and depth maps. A very efficient discriminator network then regresses the pose from the generated depth image. We argue that end-to-end learning of the ``crossed'' networks is highly beneficial for pose estimation for several reasons.  First, this architecture implicitly encodes skeleton constraints as learned from the pose data distribution. Second, the generator network effectively serves to augment the training set and improves generalization by encouraging the discovery of general representations of the observed depth data in the discriminator network. Finally, the architecture naturally allows for exploiting also unlabeled data in a semi-supervised manner. 

We learn our discriminator in a multi-task setting.
First the discriminator must be able to measure the difference between two given depth maps in the latent space.
For the generator, synthesized images from random noise are encouraged to have a desired difference to some labeled reference depth maps as measured by the discriminator. This results in a generator that produces smoother results w.r.t. the latent space. The second task of the discriminator is the standard GAN task of disambiguating real and synthesized depth maps. The posterior estimation of the hand pose, which is at the core of our method, is the third task for the discriminator. All three tasks
share the same input features, \ie the first several layers of the network and enables the posterior estimation to benefit from unlabeled and synthesized samples.

The resulting estimation framework is evaluated on 3 challenging benchmarks. Due to its simple network architecture, our method can run in real-time on the CPU and achieves results comparable or better than state-of-art with more sophisticated models. Our contributions can be summarized as follows:

\begin{itemize}
    \item We extend the GAN to a semi-supervised setting for real-valued structured prediction. Previous semi-supervised adaptations of the GAN~\cite{semiGAN,conditionalGAN,catGAN,improveGAN} have only focused on classification and are based on the fundamental assumption that the latent distribution is multi-modal with each mode corresponding to one class.

This assumption does not hold for the continuous pose regression task, since the underlying distribution of the depth map latent space does not necessarily feature multiple distinct modes.
    \item We tackle posterior estimation within a multi-task learning framework. We take advantage of the GAN to synthesize highly realistic and accurate depth maps of the articulated hand during training. Compared to a baseline which estimates the posterior directly, the multitask setting estimates more accurate poses, with the difference becoming especially prominent when training data is scarce. 
    \item  The learned generator synthesizes realistic depth maps of highly articulated hand poses under dramatic viewpoint changes while remaining well-behaved w.r.t. the latent space. Our novel distance constraint enforces smoothness in the learned latent space so that performing a random walk in the latent space corresponds to synthesizing a sequence of realistically interpolated poses and depth maps (see Fig.~\ref{fig:render}).     
\end{itemize}

\section{Related Works}
\paragraph{Deep Generative Models}
The generative adversarial network\,(GAN)~\cite{GAN} and the variational autoencoder\,(VAE)~\cite{VAE} are two recently proposed deep generative models. Typically, determining the underlying data distribution of unlabeled images can be highly challenging and inference on such distributions is highly computationally expensive and or intractable except in the simplest of cases.  GANs and VAEs provide efficient approximations, making it possible to learn tractable generative models of unlabeled images.  We provide a more detailed description in Section \ref{sec:preliminaries} and refer the reader to~\cite{GAN,VAE} for a more exhaustive treatment.

Recent works have extended the VAE~\cite{semiVAE,Sohn2015-ci,conditionalVAE} and the GAN~\cite{conditionalGAN,catGAN,semiGAN,improveGAN} from unsupervised to semi-supervised settings, though only for classification tasks. These works assume a multi-modal distribution in the latent space; while fitting for classification, this assumption does not hold for real-valued structured prediction, as is the case for hand pose estimation. Other works~\cite{draw,lapGAN,dcGAN,ssGAN,improveGAN} modify the generation model to improve synthesis. For example, the methodology in~\cite{dcGAN,improveGAN} stabilized the training process of the GAN, resulting in higher quality synthetic samples.
We use the fully convolutional network as proposed in~\cite{dcGAN} as the GAN architecture and the feature matching strategy proposed in~\cite{improveGAN}.

Since it is not possible to estimate the posterior on the GAN,~\cite{biGAN,advInfer,infoGAN} have extended the GAN to be bidirectional. Our proposed network most resembles~\cite{infoGAN}, which also formulates posterior estimation as multi-task learning.  However, instead of only estimating a subvector of the latent variable and leaving the rest as random noise as in~\cite{infoGAN}, we learn the entire posterior.
Some other works extend the GAN to cover multiple domains, and synthesize images from text~\cite{textGAN,dualGAN} or from another image domain~\cite{coupledGAN,domainGAN}. We tackle a far more challenging case of synthesizing depth maps from given poses. The synthesized depth map need to be very accurate to correspond to the given pose parameters and indeed they are, as we are even able to use synthesized images for training.

\paragraph{Hand pose estimation}
Hand pose estimation generally falls into two camps,~\ie model-based tracking and frame-wise discriminative estimation. Conventional methods need either manually designed energy functions to measure the difference between synthesized samples and observations in model-based tracking~\cite{forth-pso,msra_modelbased,chi15,Tang15,handObject,Tzionas2016,Taylor2016} or hand-crafted local~\cite{Tang2013-sf,Tang2014-lrf,Sun2015-cascade,handNormal,Tang15} or holistic~\cite{Choi_collaborative} features for discriminative estimation. 

Most recent works~\cite{nyu_hand,Oberweger_15deep,Oberweger_15feedback,deepHand,Ge_multiview,handSA,Tekin2016,Vodopivec2016,ge2017conv3D} apply convolutional neural networks\,(CNNs), and combine feature extraction and discriminative estimation into an end-to-end learning framework. Since CNNs need lots of labeled training data, semi-automatic methods have been proposed most recently~\cite{Oberweger_16,yuan2017bighand2} for accurate annotation but still take lots of efforts. On the other hand, few works have considered utilizing more easily accessible unlabeled depth maps to learn better representations. In that sense, our work resembles~\cite{Tang2013-sf} which tries to correlate unlabelled depth maps. While~\cite{Tang2013-sf} takes a discriminative approach to learn a transductive random forest, our generative approach is able to capture the distribution of unlabeled depth maps.

Our work is inspired by~\cite{Ek2007-ic,Navaratnam2007-mo}, which learned a shared manifold for observations and pose parameters based on the Gaussian process latent variable model\,(GPLVM). Another similar line of works are~\cite{Ding2016-jgp,Zhang2016-hg}, which try to learn a shared latent space between pose and gait also based on GPLVM. The GPLVM is a non-parametric model, whereas our generative model is in the format of neural network, which makes it possible to learn the generative models together with the posterior estimation in an end-to-end manner.

\section{Preliminaries}
\label{sec:preliminaries}

\begin{figure*}
\centering
\includegraphics[width=\textwidth]{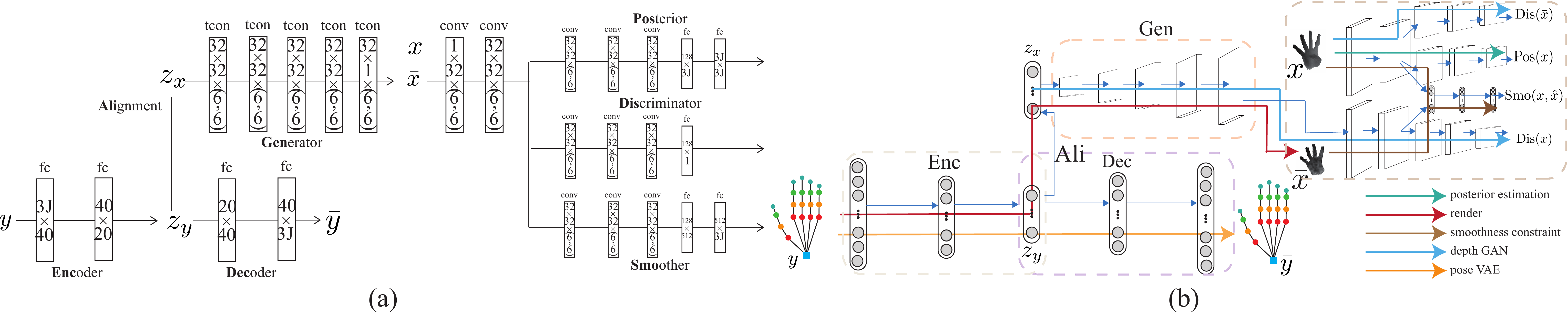}
\caption{{\bf Overview of the proposed system.} 
(a) shows the network architecture and a sketch of the variable relationships. \textbf{fc} stands for fully connected layers, \textbf{tcon} stands for transposed convlutional layers with dialation factor of 2, and \textbf{conv} stands for convolutional layers with stride of 2. Numbers inside the boxes denote the parameter size.
(b) depicts the data flows within the network used in our work. Arrows with different colours indicate data flows associated with a specific task as shown in the legend. See Section \ref{sec:overview} for details. The figure is best viewed in colour.
}
\label{fig:framework}
\end{figure*}

Let $o$ represent some observation (either the hand pose or the depth map).  We wish to estimate a prior $p(o)$ by modeling the generation process of $o$ by sampling some $z$ from an arbitrary low-dimensional distribution $p(z)$ as $p(o) = \int_{z} p(o|z)p(z)dz$. Fitting $p(o)$ directly is intractable and usually involves expensive inference. We therefore approximate $p(o)$ using two recently developed and very powerful deep generative models: the variational autoencoder\,(VAE) and the generative adversarial network\,(GAN). 

In the remainder of this section, we provide a brief introduction of the VAE and GAN which we use to model the prior of hand poses and depth maps. Notation-wise, we refer to a given depth map as $x$ and a hand pose as $y$. We denote the latent variable as $z$ and further distinguish as $z_x$ and $z_y$ indicate the latent depth map and pose respectively when the distinction is necessary.  $\bar{x}$ refers to the synthesized depth map from GAN generator and $\bar{y}$ to the reconstructed pose parameter from VAE decoder.

\subsection{Pose Variational Autoencoder (Pose VAE)}
A VAE comprises an \textit{encoder} which estimates the posterior of latent variable and a \textit{decoder} generates sample from latent variable as follows,
\begin{equation}
z_y \sim \enc(y) = q(z_y|y),~
\bar{y} \sim \dec(z_y) = p(y|z_y).
\end{equation}

The VAE regularizes the encoder by imposing a prior over the latent distribution on $p(z_y)$ while at the same time reconstructing $\bar{y}$ to be as close as possible to the original $y$. Typically, a Gaussian prior is used, \ie $z_y\sim \mathcal{N}(0,I)$, and is incorporated into the loss as the Kullback-Leibler divergence $D_{\text{KL}}$ between the encoded distribution $q(z_y | y)$ and the prior $p(z_y)$. The VAE loss is then the sum of the reconstruction error and latent prior:
\begin{equation}
\loss_{\vae} = \loss_{\recons}^{pose} + \loss_{\mbox{prior}},
\end{equation}
where 
\begin{equation}
\loss_{\recons}^{pose}\! =\! -\mathbb{E}_{q(z_y|y)}[\log p(y|z_y)] 
\end{equation}
and 
\begin{equation}
\loss_{\mbox{prior}} = D_{\text{KL}}(q(z_y|y)||p(z_y)).
\end{equation}

We use the VAE to model a prior distribution on hand pose configurations. The encoder-decoder structure allows us to learn a mapping from high dimensional hand poses to a low-dimensional representation while ensuring a high reconstruction accuracy through the decoder. Furthermore, the constraint on the latent distribution simplifies the learning of a shared latent space of between the depth map and the pose (see Section~\ref{sec:shared} for details).

\subsection{Depth Map Generative Adversarial Network (Depth GAN)}
A GAN consists of a \textit{generator} and a \textit{discriminator}. The generator synthesizes samples by mapping a random noise sample $z_x$, from an arbitrary distribution, to a sample in the data space $\bar{x}$.
The discriminator tries to distinguish between real data samples $x$ and synthesized samples $\bar{x}$ from the generator. The loss function for the GAN can be formulated as a binary entropy loss as follows:
\begin{equation}
\loss_{\gan} = \log(\dis(x)) + \log(1-\dis(\gen(z_x))),
\end{equation}
where $\dis(x)$ is the discriminator output and is a measure of the probability of $x$ being a real data sample.
Training alternates between minimizing $\loss_{\gan}$ w.r.t. parameters of the generator while maximizing $\loss_{\gan}$ w.r.t. parameters of the discriminator.
The generator tries to minimize the loss to generate more realistic samples to fool the discriminator while the discriminator tries to maximize the loss. 

The GAN does not explicitly model reconstruction loss of the generator; instead, network parameters are updated by back-propagating gradients only from the discriminator. This effectively avoids pixel-wise loss functions that tend to produce overly smoothed results and enables realistic modeling of noise as present in the training set. 
The GAN can therefore generate depth images with high realism and learn latent representations with linear semantics, \ie simple arithmetic operations in the latent space can result in semantic transformations in the data space~\cite{dcGAN,improveGAN}.  As such, the GAN is well-suited to model the generation process of depth maps and can be used, together with the shared latent space, for synthesizing samples to augment the training corpus.  
In this work, we adopt a deep convolutional GAN network architecture of~\cite{dcGAN} and a feature matching strategy~\cite{improveGAN} for stable and fast-converging training. The noise is sampled from a uniform distribution as $z_x\sim \mathcal{U}(-1,1)$.

\section{Method}
\label{sec:method}

\subsection{System Overview - Crossing Nets}
\label{sec:overview}
We formulate hand pose estimation as a statistical learning problem: given a corpus of depth maps, we aim to learn a posterior distribution over the corresponding hand poses.  We approach this by combining two generative neural networks, one for pose, and one for depth appearance.  First, we pre-train each network separately to capture statistics of the individual domains.  We then learn a mapping between the two latent spaces $z_x$ and $z_y$. The complete network is then further trained end-to-end for the pose estimation task. Fig.~\ref{fig:framework} gives an overview of our architecture.

In Fig.~\ref{fig:framework}, the blue and yellow routes represent the forward paths of the VAE and the GAN for pose and depth map respectively. The blue route,~\ie the render route, links the VAE and the GAN together through the mapping~\textit{Ali}. Given any pose, the data is forwarded through the blue route and the network can synthesize a depth map with the corresponding pose. Details of training the render route is given in Section~\ref{sec:shared}. The green route estimates the posterior of shared latent variable given the depth map, while the brown route places a smoothness constraint on the generator of GAN. Both the green and the brown 
routes share the parameters with the discriminator of GAN, with details described in Section~\ref{sec:posterior}. 

Neglecting sensor noise, we assume that there is a one-to-one mapping between the depth map and the hand pose for the free moving hand. As such, we can arbitrarily choose either the pose or the depth map latent space as the reference shared space and then learn a mapping to the other latent space to link the two generative models together. We show how this mapping is learned in Section \ref{sec:shared}.

To prevent from over-fitting, we formulate the posterior estimation as a multi-task learning in which all tasks share the first several convolutional layers. In addition to latent variable regression or \textit{posterior task}, we also consider a \textit{smoothness task} and the \textit{GAN task}. By jointly training the generator and discriminator, as explained in Section \ref{sec:posterior}, our method can benefit from the unlabeled samples as well as synthesized samples from the generator.

\subsection{Learning the Shared Latent Space}
\label{sec:shared}
It is not possible with the machinery of the depth GAN alone to estimate the latent variable posterior. As such, we must first learn a mapping from one latent space to the other. We choose the latent space of hand pose parameter as the reference space and learn a mapping to the depth map latent space, \ie $z_y = \ali(z_x)$. Note that we do not have training pairs of corresponding $(z_x, z_y)$.  What we do have, however, are corresponding pairs $(x,y)$, so it is possible instead to compare observed depth images $x$ with synthesized images $\bar{x}$ that are projected to $z_y$ and then mapped to $z_x$.  
As such, we introduce a proxy loss $\loss_{\recons}$, based on the reconstruction error of the rendered depth map given a latent input $z_{y}^{(i)} = \enc(y^{(i)})$ which is mapped to the GAN latent space:   
\begin{equation}
\loss_{\recons} = \frac{1}{N}\sum_i^N \max(\|x^{(i)}-\gen(\ali(z_{y}^{(i)}))\|^2,~\tau),
\label{equ:recons}
\end{equation}

We model $\ali(\cdot)$ as a single fully connected neuron with a $\tanh$ activation. The forward pass corresponds to the purple route in Fig.~\ref{fig:framework}. Similar to the golden energy used in~\cite{chi15}, we use a clipped mean squared error for our loss function, to remain robust to depth sensor noise. Since the depth map is normalized to $[-1,1]$, we set the clip threshold $\tau = 1$. 

Parameters of the mapping $\theta_{\ali}$ are  optimized through back-propagation. Since both the pose VAE and the depth GAN are able to learn low-dimensional representations (our $z_x$ and $z_y$ are both 23 dimensions respectively), we are able to fit the alignment and generate realistic samples with very few labeled $(x,y)$ pairs.  

After the mapping $\ali(\cdot)$ is learned, any point in the latent pose space can then be projected into both a hand pose (through the pose VAE) or into a corresponding depth map (through the depth map GAN). We can therefore regard the two latent spaces synonymously as a common shared latent pose. The composite function  $\gen(\ali(\cdot))$ acts as the new generator for the depth latent space.

Since we impose a Gaussian prior $\mathcal{N}(0, I)$ on $z_y$, ideally, any random noise sampled from the standard normal distribution can be mapped both to a hand pose or a corresponding depth map. Note that $\ali(\cdot)$ is implicitly learning a mapping from a normal distribution ($z_y$) to a uniform distribution ($z_x$).

\subsection{Learning the posterior of shared latent variable}
\label{sec:posterior}
There are three types of data we can use to learn the latent posterior: labeled samples $(X_l, Y_l)$, synthesized samples from random noise $(Z_r, \bar{X}_r = \gen(\ali(Z_r)))$ and unlabeled depth maps $X_u$. In this section, we overload our notation and use the capital letters to indicate mini-batch data matrices of $N$ columns, where each column vector is a sample. For any given matrix $A$, we use $\|A\|_{*}$ to indicate the sum of Euclidean norms for each column vector, \ie  $\|A\|_{*} = \sum_{j=1}^n(\sum_{i=1}^m |a_{ij}|^2)^{\frac{1}{2}}$.

Although it is theoretically sufficient to use only $(X_l,Y_l)$ pairs for learning the posterior, one does not fully exploit the learned priors from the depth GAN. To allow the posterior estimate to benefit from also synthesized and unlabeled samples and therefore increase generalization power, we add two more tasks, \ie a smoothness task and a GAN disambiguation task. All three tasks share the first several convolutional layers, taking synthesized and unlabeled samples as input to exploit the benefits of the depth GAN. 

To encourage the generator to synthesize more accurate and realistic samples, parameters $\theta_{\ali}$ and $\theta_{\gen}$ of the composite generation function $\gen(\ali(\cdot))$ are updated together with the aforementioned multitasks. For simplicity, we use \textit{generator} to indicate the composite function of $\gen(\ali(\cdot))$ which takes noise from the shared latent space as input and generates a depth map.  We use \textit{discriminator} to indicate the multitask learning as a whole, taking depth maps as input. In each iteration, both the generator and the discriminator are updated jointly. The discriminator is updated with labeled, unlabeled and synthesized samples; at the same time, the generator is updated through back-propagated gradients from discriminator. The joint update ensures that the generator synthesizes progressively more realistic samples for the discriminator. We define the joint generator and discriminator loss as
\begin{align}
\loss_{G} &= \loss_{\recons} + \loss_{\smo} - \loss_{\gan},\label{equ:lossG}\\
\loss_{D} &= \loss_{\pos} + \loss_{\smo} + \loss_{\gan},\label{equ:lossD}
\end{align}
where $\loss_{G}$ represents the generator loss and $\loss_{D}$ the discriminator loss.

\textbf{Smoothness task.} To encourage the underlying latent space to be smooth, we define a $\loss_{\smo}$ for both the generator and the discriminator.
Given two depth maps $x_1,~x_2$ and their corresponding underlying latent variables $z_1,~z_2$, the smoothness $\smo(x_1,x_2)$ task takes $x_1$ and $x_2$ as input and estimates the corresponding latent variable difference $z_1 - z_2$. The estimated difference is then compared to the actual difference. 
To make $\loss_{\smo}$ regularize both discriminator and generator, we substitute one of the latent-observation pairs with random noise $z_r$ and the corresponding synthesized image $\bar{x}_r$, as indicated by $d_{\mbox{comb}}$ in Eq.~\ref{equ:gen}.  At the same time, we want the projected $z_l$ of the labeled sample to synthesize into an image as close as possible to the original so we add the term $d_{\mbox{self}}$, resulting in the following smoothness loss:
\begin{equation}
\begin{split}
\loss_{\smo} =& d_{\mbox{comb}} + d_{\mbox{self}}\\
			 =& \frac{1}{N}\|\smo(\bar{X}_r, X_l) - (Z_r-Z_l) \|^2_{*}\\
              & + \frac{1}{N}\|\smo(\bar{X}_l, X_l)\|^2_{*}.
\end{split}
\label{equ:gen}
\end{equation}
Here, $X_l$ is a set of labeled depth maps, $Z_l = \enc({Y_l})$ is their corresponding latent variable and $\bar{X}_l = \gen(\ali(Z_l))$ the depth maps reconstructed through the generator.
$\bar{X}_l$ is also compared to the depth maps  $\bar{X}_r = \gen(\ali(Z_r))$, synthesized from a set of random noise vectors $Z_r$  in the latent space. 
In practice, the $\smo(\cdot,\cdot)$ operation is implemented as a Siamese network as depicted in Fig~\ref{fig:framework}.

\textbf{GAN task.} Although disambiguating real from synthetic samples is not directly linked to posterior estimation, it has been shown in several previous works~\cite{ssGAN,semiGAN,conditionalGAN,dcGAN} having such a loss encourages the hidden activations of the discriminator to learn, as the name implies, inherently discriminative features without additional supervision. We therefore add the following GAN loss term 
\begin{equation}
\loss_{\gan} = \frac{1}{N} \|\log(\dis(X)) + \log(1-\dis(\gen(Z)))\|_{*}^2,
\end{equation}
where $X = X_l \cup X_u$ is the union of labeled and unlabeled depth maps and $Z = Z_l \cup Z_r$ is the union of synthesized depth maps from latent variables of labeled samples and randomly sampled ones from prior distribution.

\textbf{Posterior task.} Given an input depth map, we formulate a loss for the shared latent variable posterior as
\begin{equation}
\loss_{\pos} = \frac{1}{N} \|\pos(X_l) - Z_l \|_{*}^2,
\end{equation}
where $\pos(X)$  maps the training set of depth maps $X$ to the corresponding shared latent variable vector $Z$. $Z_l$ is the set of target positions in the latent space, as obtained by the VAE.

\textbf{Multi-task Training.} 
We additively combine the three loss functions into a single loss function, using equal weights. In each training iteration, both the generator and the discriminator network parameters are updated once. The detailed training procedure is shown in Algorithm ~\ref{alg:training}.

\begin{algorithm}
    \small
    {
        \caption{Training the posterior via multitask learning}
        \label{alg:training}
        \begin{algorithmic}[1]
        \INIT 	$\theta_{\ali},\theta_{\gen},\theta_{\dis}~~\gets~~$ initialized through pretraining
		\INIT $\theta_{\smo},\theta_{\pos}~~\gets~~$ randomly initialized
        \State $\theta_{G} := \theta_{\ali} \cup \theta_{\gen}$
        \State $\theta_{D} := \theta_{\smo}\cup \theta_{\pos} \cup \theta_{\dis}$
        \For{number of training epoch}
        \State $X_l,Y_l$ $\gets$ random minibatch labeled pairs
        \State $X_u$ $\gets$ random minibatch unlabeled depth map
        \State $Z_r$ $\gets$ random noises sampled from $p(z)$
        \State $Z_l, \bar{X}_l, \bar{X}_r~\gets~\enc(Y_l), \gen(\ali(Z_l)), \gen(\ali(Z_r))$
        \State $X_1, X_2, Z_1, Z_2~~\gets$ random equal split of $X$ and $Z$
        \State $X, Z \gets X_l \cup X_u, Z_l \cup Z_r$
        \State $d_{\mbox{comb}} := \frac{1}{N}\|\smo(\bar{X}_r, X_l) - (Z_r-Z_l) \|^2_{*}$
        \State $d_{\mbox{self}} :=  \frac{1}{N}\|\smo(\bar{X}_l, X_l)\|^2_{*}$
        \State $\loss_{\smo} \gets d_{comb} + d_{self}$
        \State $\loss_{\recons} \gets \|\max(\|X_l-\bar{X}\|,~ \tau)\|_{*}^2$
        \State $\loss_{\pos} \gets \|\pos(X_l) - Z_l \|_{*}^2$
        \State $\loss_{\gan} \gets \frac{1}{N} \|\log(\dis(X)) + \log(1-\dis(\gen(Z)))\|_{*}^2$
 		\State $\theta_{D} \gets \theta_{D} - \nabla_{\theta_{D}}(\loss_{\pos} + \loss_{\smo} - \loss_{\gan})$
        \State $\theta_{G} \gets \theta_{G} - \nabla_{\theta_{G}}(\loss_{\recons} + \loss_{\smo} + \loss_{\gan})$
        \EndFor
        \end{algorithmic}
    }
\end{algorithm}

\subsection{Implementation details}
The first 2 convolutional layers of the discriminator network is shared by the three tasks (smoothness task, GAN task and posterior estimation). To stabilize training, we use batch normalization on every hidden layer.
Instead of sampling noise from the prior distribution, we generate random noise as the convex 
combination with random weights from the labeled latent variables. We use the Adam~\cite{Kingma2014-adam} method to update network parameters. To make the generator and discriminator more robust, we injected random Gaussian noise with 0.05 standard deviation to the latent variable after 
VAE encoder $\enc(\cdot)$ during training.  We set the learning rate as 0.001 and train the complete network for 100 epochs. It takes about 10 hours for training with around 70k samples on one Nvidia TITAN X GPU. 
 
\section{Experiments}
We performed experiments on 3 publicly available datasets. As each dataset has its own set of challenges, we briefly summarize their characteristics in Table~\ref{Tab:datasets}. NYU is quite noisy and has a wide range of poses with continuous movements, while  MSRA is limited to 17 gestures but has many viewpoint changes. 
ICVL has large discrepancies between training and testing; test sequences are with fast and abrupt finger movements whereas training sequences have continuous palm movement with little finger movement.

While we estimate all 36 annotated joints on NYU, we only evaluate on a subset of 14 joints as in~\cite{nyu_hand,Oberweger_15deep,Oberweger_15feedback} to make a fair comparison.

\begin{table}[b!]
\footnotesize
\begin{tabular}{|c|c|c|c|}
\hline
Dataset & Depth Sensor & Train/Test & Noise \\ 
\hline
NYU~\cite{nyu_hand} & PrimeSense & 72.7k / 8.2k & high \\ 
MSRA~\cite{Sun2015-cascade} & Intel RealSense & 76.5k, 9 users / leave-user-out & low \\
ICVL~\cite{Tang2014-lrf} & Intel RealSense  & 20k (160k) / 1.6k & low \\ 
\hline
\end{tabular}
\caption{{\bf Hand pose estimation benchmarks.}}\label{Tab:datasets}
\end{table}

\begin{figure*}
    \centering
    \includegraphics[width=\linewidth]{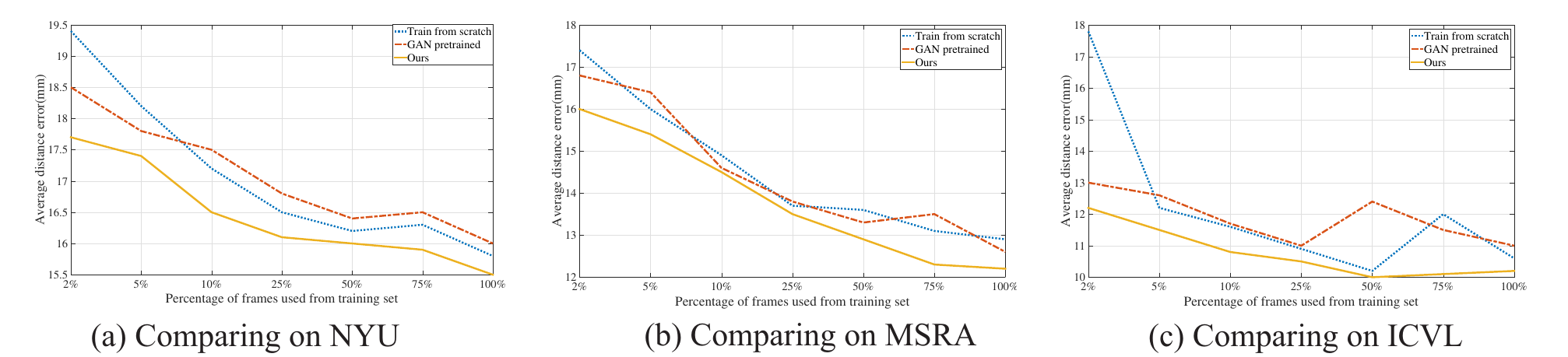}
    \label{fig:semi}
    \caption{{\bf Semi-supervised learning.} Comparison of our approach and two baseline methods when using $m\%$ of frames from the training set as labeled data, and discarting the labels of the other images.}
    
    \vspace{1.5cm}    
    
    \includegraphics[width=\linewidth]{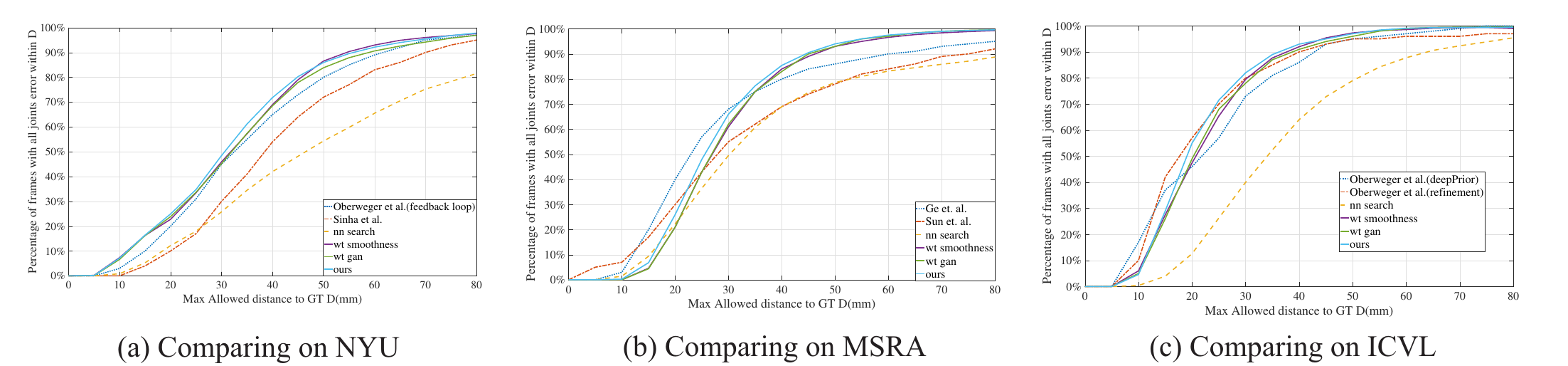}
    \caption{{\bf Comparison of our approach with state-of-the-art methods.} We compare our approach with of previous method on three challenging datasets.}
    \label{fig:comparison}
    
    \vspace{1.5cm}

    \includegraphics[width=\linewidth]{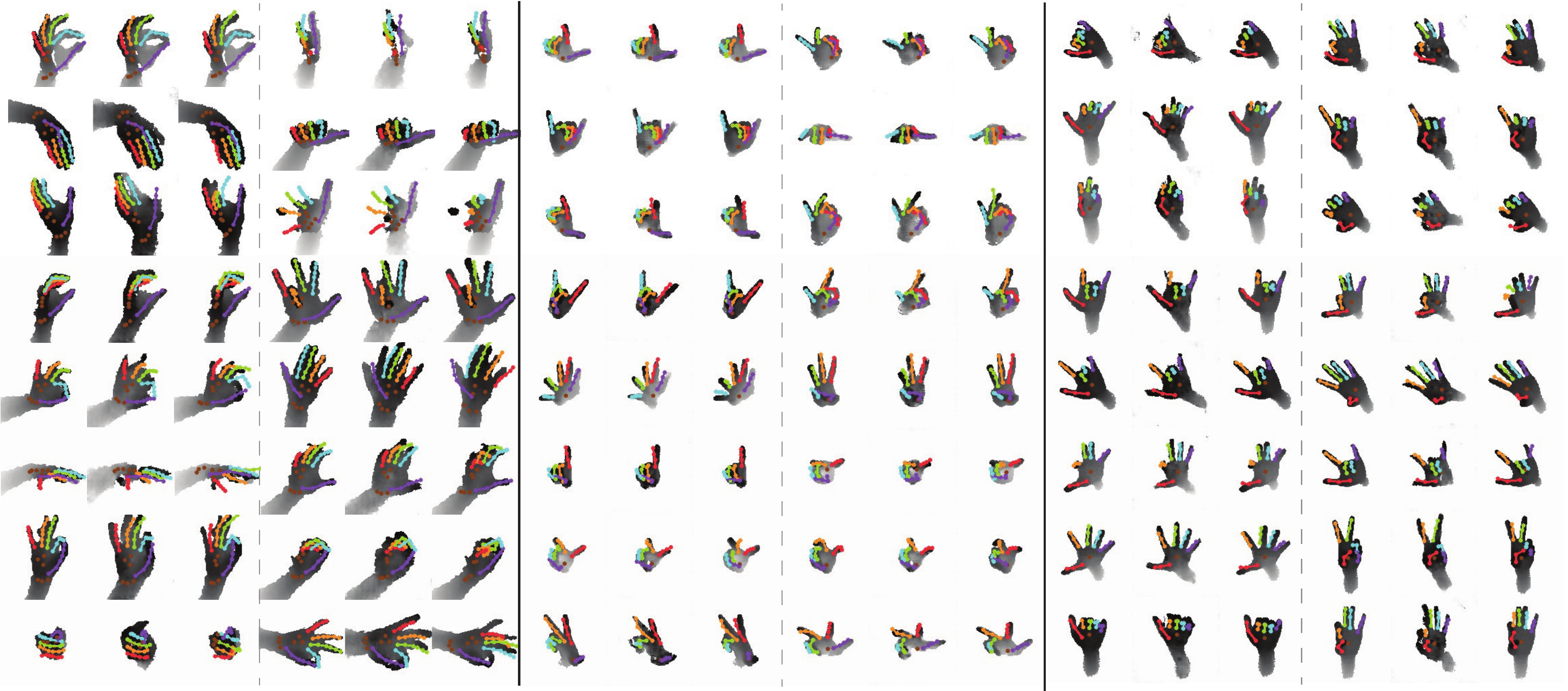}
    \caption{{\bf Qualitative hand pose estimation results.} Left: NYU\cite{nyu_hand}, middle: MSRA\cite{Sun2015-cascade}, right: ICVL\cite{Tang2014-lrf}. For each sample triplet, left is ground-truth, middle is reconstructed depth map and pose from shared latent space, right is estimated result.}
    \label{fig:qual}
\end{figure*}

We quantitatively evaluate our method with two metrics: mean joint error (in mm) averaged over all joints and all frames, and percentage of frames in which all joints are below a certain threshold~\cite{Vitruvian}.  Qualitative results are shown in Fig.~\ref{fig:qual} for estimation results and Fig.~\ref{fig:render} for the synthesized images from the neural network. We encourage the reader to watch the supplementary videos for a closer qualitative look. The networks were implemented with the Theano package\cite{theano}; on an Intel 3.40 GHz i7 machine, the average run time is 11\.ms per image (90.9\.FPS). 

\subsection{Semi-supervised learning}\label{sec:ssl_res}

To explore how our method performs in the semi-supervised setting, we uniformly sample $m\%$ of frames from the training set as labeled data and use the remaining frames unlabeled. 
We then vary $m$ from 2\% to 100\% and evaluate the mean joint error averaged over all joints and all frames.  We compare against two baseline posterior estimation methods: one network trained from scratch (using randomly initialized parameters),
and one network where the first two convolutional layers are initialized with parameters from a GAN pretrained on the entire training set. 

Unsurprisingly, when $m\!=\!2\%$, both the GAN-pretrained baseline and our semi-supervised setup achieve better results than training from scratch. This validates that our depth GAN is effective at learning good representations in an unsupervised way. However, the GAN-pretrained method does not give better results than training from scratch when $m \geq 5\%$. A more unexpected finding was that using more training samples did not lead to a monotonous decrease in the average joint error on both baselines. We attribute this to two causes.  First, the labeled frames are uniformly sampled.  Since in all three datasets there is slow continuous movement, there is a high correlation between the frames; a 5\% sampling 
may already cover a large portion of distinct hand poses 
and more samples does not add substantially more information. Secondly, as we evaluate based on the number of training epochs, 
having more training samples effectively results in more gradient updates and may lead the network to over-fitting.  Nevertheless, our method always outperforms the two baselines, showing that using synthesized and unlabeled samples does help with network generalization and preventing overfitting.

\subsection{Contribution of multi-task learning}
The comparison against the baselines described in Section~\ref{sec:ssl_res} demonstrates that our multi-task learning outperforms direct posterior estimation, both in a semi-supervised and fully supervised setting. To investigate the independent contributions of each energy term in detail, we introduce two more baselines: one trained without the smoothness loss $\loss_{\smo}$ and another one without the GAN loss $\loss_{\gan}$. The results (plotted as solid lines in Fig.~\ref{fig:comparison}) evince that our multi-task approach consistently outperforms both baselines, validating the effectiveness of $\loss_{\smo}$ and $\loss_{\gan}$ terms.

\subsection{Comparison with State-of-the-Art}
We compare the accuracy of our method with 6 previous state-of-art methods.  In general, our results show that our method is either on par with competing methods or even outperforming them. Compared to hierarchical methods\cite{Sun2015-cascade,Oberweger_15deep,Ge_multiview}, our results are slightly worse at low error thresholds. This demonstrates a general pattern: holistic methods that estimate the hand as a whole tend to be more robust but not very accurate at estimating the finger pose. Hierarchical methods on the other hand estimate the finger pose conditioned on the estimated palm pose and are therefore more accurate, but are also sensitive to the noisy estimation of palm pose. Meanwhile, inspired by~\cite{Rogez2015-vy,review15} we also compare against a nearest neighbour searching based baseline(indicated as nn-search in Fig.~\ref{fig:comparison}), in which PCA is used reduce the input depth map into a 512 dimensional feature vector followed by nearest neighbour searching. Given the training and testing samples are similar, the nn-search baseline works reasonably well as on MSRA and vice versa on NYU and ICVL. 

On NYU, we compare with Sinha \textit{et al.}~\cite{deepHand} and Oberweger \textit{et al.}(feedback loop)~\cite{Oberweger_15feedback}. As shown in Fig.~\ref{fig:comparison} (a), we outperform~\cite{deepHand,Oberweger_15feedback} by a large margin. 

On MSRA, we compare with Ge~\textit{et al.}~\cite{Ge_multiview} and Sun \textit{et al.}~\cite{Sun2015-cascade}. 
Since our approach is holistic, it is not as accurate as the hierarchical methods of~\cite{Ge_multiview,Sun2015-cascade} on error threshold from 10-30mm.  However, we outperform these two when the error threshold is larger than 35mm, which we attribute to our method being more robust to large viewpoint changes.

On ICVL, we compare with thet two variations (deepPrior) and (refinement) of Oberweger~\textit{et al.}\cite{Oberweger_15deep}. We outperform (deepPrior) when the error threshold is $\geq20$mm with a large margin. Compared to the much more sophisticated (refinement) variation, which
refines the estimate of each joint via a cascaded network,
our method is better by 2\% with error thresholds when error threshold is $\geq30$mm.

\section{Conclusion}
In this paper, we propose a hand pose estimation method by estimating the posterior of the shared latent space of depth map and hand pose parameters. 
We formulate the problem as a multi-task learning problem on a network architecture that crosses two deep generative networks: a variational auto encoder\,(VAE) for hand poses and a generative adversarial network\,(GAN) for modeling the distributions of depth images. By learning a mapping between the two latent spaces, we can train the complete network end-to-end. 
In our experiments we demonstrate that this has a number of advantages: we can exploit the generalization properties of the GAN as well as the pose constraints implicitly learned by the VAE to improve discriminative pose estimation. Moreover, our architecture naturally allows for learning from unlabeled data, which is very valuable for the problem of hand pose estimation, where annotated training data is sparse. Our approach therefore extends the semi-supervised setting of GAN to making real valued structured predictions. 
We evaluated our method on 3 publicly available datasets and demonstrate that our approach consistently achieves better performance over previous state-of-art methods. Due to a very efficient design of the discriminator network our approach is capable of running in real-time on the CPU.

\noindent \textbf{Acknowledgment} The authors gratefully acknowledge support by armasuisse, KTI project with Faswhell and Chinese Scholarship Council.

{\small
\bibliographystyle{ieee}
\bibliography{egbib}
}

\end{document}